\documentclass{edm_article}
\usepackage[utf8]{inputenc}
\usepackage[T1]{fontenc}

\usepackage{amsmath}
\usepackage{amssymb}

\usepackage{enumitem}
\setenumerate{itemsep=1pt,topsep=1pt}

\usepackage{graphicx}

\usepackage[numbers]{natbib}

\usepackage{tabularx}

\usepackage{hyperref}
\usepackage{cleveref}

\hypersetup{
	pdfauthor={Anaïs Tack, Chris Piech},
	pdftitle={The AI Teacher Test: Measuring the Pedagogical Ability of Blender and GPT-3 in Educational Dialogues},
	pdfkeywords={student-teacher dialogue, conversational agents, chatbots, Blender, GPT-3, evaluation methods, pairwise comparisons, Bayesian Bradley-Terry model},
	pdfsubject={},
	pdfcreator={}, 
	pdflang={English}}

 \numberofauthors{2}
\author{
	\alignauthor
	Anaïs Tack\\
	\affaddr{Stanford University}\\
	\email{atack@cs.stanford.edu}
	\alignauthor 
	Chris Piech\\
	\affaddr{Stanford University}\\
	\email{piech@cs.stanford.edu}
}
\title{The AI Teacher Test: Measuring the Pedagogical Ability of Blender and GPT-3 in Educational Dialogues}
\begin{document}

\maketitle
\begin{abstract}
How can we test whether state-of-the-art generative models, such as Blender and GPT-3, are good AI teachers, capable of replying to a student in an educational dialogue?
Designing an AI teacher test is challenging: although evaluation methods are much-needed, there is no off-the-shelf solution to measuring pedagogical ability.
This paper reports on a first attempt at an AI teacher test.
We built a solution around the insight that you can run conversational agents in parallel to human teachers in real-world dialogues, simulate how different agents would respond to a student, and compare these counterpart responses in terms of three abilities: speak like a teacher, understand a student, help a student.
Our method builds on the reliability of comparative judgments in education and uses a probabilistic model and Bayesian sampling to infer estimates of pedagogical ability.
We find that, even though conversational agents (Blender in particular) perform well on conversational uptake, they are quantifiably worse than real teachers on several pedagogical dimensions, especially with regard to helpfulness (Blender: \(\Delta \, \text{ability} = -0.75\); GPT-3: \(\Delta \, \text{ability} = -0.93\)).
\end{abstract}


\section{Introduction}
\label{sec:orge8b694c}
Conversational agents (or chatbots) offer promising opportunities for education. They can fulfill various roles (such as intelligent tutors and service-oriented assistants) and pursue different objectives (e.g., improving student skills, boosting student motivation, and increasing instructional efficiency) \cite{wollny_are_2021}. Among all of these different vocations of an educational chatbot, the most prevalent one is the AI teacher helping a student with skill improvement and providing more opportunities to practice. Some recent meta-analyses have even reported a significant effect of chatbots on skill improvement, for example in language learning \cite{bibauw_dialogue_2022}.
What is more, current advances in AI and natural language processing have led to the development of conversational agents that are founded on more powerful generative language models. Blender \cite{roller_recipes_2020}, for instance, is a state-of-the-art open-domain chatbot trained to blend skills such as being empathetic and knowledgeable \cite{smith_can_2020}, which are undeniably important characteristics of a good AI teacher.
Furthermore, the current state-of-the-art in natural language generation is GPT-3 \cite{brown_language_2020}, a 175B-parameter model that is able to multitask different language generation skills (such as conversation). The astonishing power of GPT-3 is that it can perform these skills with few-shot in-context learning, merely from seeing a short prompt describing the task at hand (e.g., \emph{The following is a conversation with an AI assistant.}). Emergent models such as GPT-3 have been described as \emph{foundation models} since they serve as the ``common basis from which many task-specific models are built via adaptation'' \citep[p.7]{bommasani_opportunities_2021}.

Despite these promising opportunities, the use of powerful generative models as a foundation for downstream tasks also presents several crucial challenges. In the educational domain in particular, it is important to ascertain whether that foundation is solid or flimsy. \citet[pp.67-72]{bommasani_opportunities_2021} stressed that, if we want to put these models into practice as AI teachers, it is imperative to determine whether they can (a) speak to students like a teacher, (b) understand students, and (c) help students improve their understanding. Consequently, there is a critical need to establish good evaluation methods of AI teachers. This is a hard problem because there is no off-the-shelf and universal solution to measuring teaching ability and effectiveness.

Therefore, we took on the challenge of designing an AI teacher test and conducted a pilot study. We ran Blender and GPT-3 in parallel to human teachers in language and mathematics educational dialogues, observed how they responded to a student, and compared these counterpart responses in terms of pedagogical ability. The major contributions of this work are as follows:
\begin{enumerate}
\item We pose the AI Teacher Test Challenge.
\item We implement a human-in-the-loop pairwise comparison test as a first attempt at an AI Teacher Test.
\item Our results show quantitatively how far conversational agents, particularly Blender and GPT-3, are behind human teachers in terms of pedagogical ability, despite them performing well on conversational uptake.
\end{enumerate}
Our solution has several strengths: (1) it leverages the proven reliability of comparative judgments in education \cite{heldsinger_using_2010,lesterhuis_comparative_2017}, (2) it incorporates a Bayesian sampling method that allows us to attribute an ability score to a teacher response, whilst ensuring normality and providing a belief in our estimates, and (3) it produces scores and ranks that could be used to develop autonomous methods.
We open-source our work, code, and data.\footnote{\url{https://github.com/anaistack/ai-teacher-test}}

\section{The AI Teacher Test Challenge}
\label{sec--ai-teacher-test-challenge}
\begin{figure}[t]
\centering

\includegraphics[width=\linewidth]{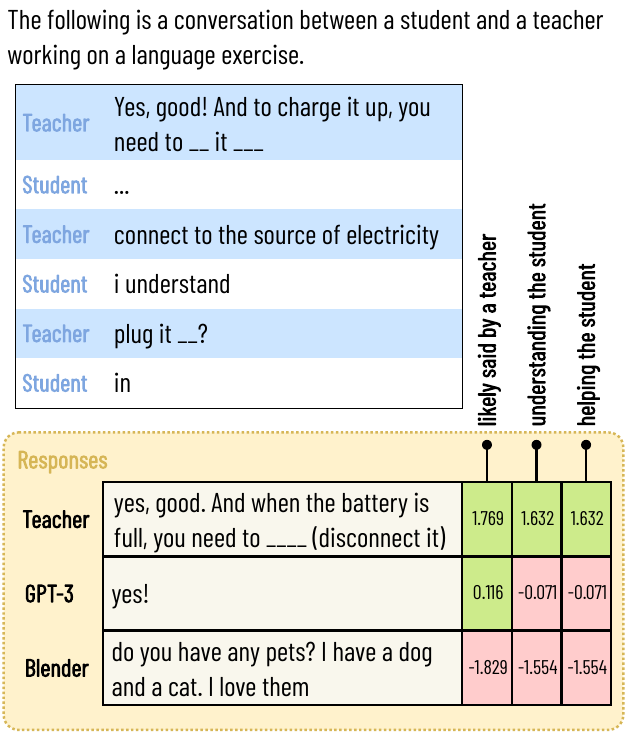}
\caption{Illustration of the AI Teacher Test Challenge: Estimates of Pedagogical Ability and Rankings of Human and AI Teachers Replying to a Student in an Educational Dialogue}
\label{fig--example-dialogue-ability}
\end{figure}

Consider the following scenario, which is illustrated in \Cref{fig--example-dialogue-ability}.
Two agents, a student and a teacher, are interacting in an educational setting.
The student is working to improve a specific skill (e.g., the use of phrasal verbs) in a given domain (e.g., English language).
The teacher could be either a human agent or an artificial agent who is helping the student with improving this skill.
The student and teacher each take turns, resulting in a sequence of student-teacher dialogic pairs.
This student-teacher dialogue is open-ended: for a given student utterance, there exists a variety of ways in which we could imagine a teacher agent to respond. For example,  \Cref{fig--example-dialogue-ability} shows three possible replies to a student utterance: the actual teacher's response and two completions that were automatically generated from a state-of-the-art language model. It is clear to see that, in the space of possible replies, not all responses will be equally preferable. Some responses may be more characteristic of a teacher, some may be taking up more from the student's utterance, and some may be more helpful.
In this scenario, we are interested in the following challenge: given a space of possible responses (either human or artificially generated), evaluate a reply in terms of pedagogical ability and estimate this score relative to other replies.

%

\subsection{Desiderata}
\label{sec:org6bdb65c}

We think that a good AI teacher test should at least account for the following aspects.
Firstly, the test should be able to evaluate a teacher agent's response in context.
At minimum, the test should consider the preceding student utterance. Additionally, the test could also take into consideration the entire preceding dialogue and surrounding educational setting.
Secondly, the test should be able to score the agent's response with respect to several pedagogical abilities.
Following \citet[pp.67-72]{bommasani_opportunities_2021}, we believe that the test should consider the following three abilities: whether the agent can speak like a teacher, understand the student, and help the student.
Finally, the test should also be able to consider other possibilities (which may be better or worse) and rank the teacher's response in comparison to these. In this way, the test could also be used to suggest one or more ways in which a response could be enhanced in terms of the three abilities listed above.

Unfortunately, standard methods of evaluating automatically generated language and conversational agents do not meet our desiderata. Perplexity, for example, measures how well a generative model is able to sample a given response from its probability distribution. However, it does not consider the preceding utterance (desideratum \#1). Other metrics such as BLEU and F1 score measure the n-gram overlap between a generated response and a correct response. By contrast, our test is open-ended (see above) and does not presuppose the existence of a correct response. Recently, \citet{pillutla_mauve_2021} introduced MAUVE, an evaluation metric for open-ended language generation. Because this metric uses the Kullback–Leibler divergence, it cannot be used to compare two specific language utterances (desideratum \#3). Most importantly, none of these methods meet our second desideratum, which is to score an agent's response with respect to several pedagogical abilities.

\subsection{Related Work}
\label{sec--related-work}
We can gain insight into measuring pedagogical ability from prior work into assessing human teachers.
Educational research is abundant in methods for evaluating teacher effectiveness, ranging from teacher self-reports and interviews to classroom observations, student evaluation surveys, and tests of student achievement \cite{goe_approaches_2008,muijs_measuring_2006}. However, not all of these methods seem easily applicable to assessing AI teachers. It is obvious that evaluating AI teacher effectiveness from self-reports and interviews would be a difficult thing to do. We could, however, resort to systematic observations of AI teachers, human evaluation surveys, and measures of student outcome.

Other studies have focused on the possibility of measuring ability in teacher language.
\citet{demszky_measuring_2021}, for instance, examined several ways of determining how well a teacher replies to a student in student-teacher interactions. Their data comprised 2,246 student-teacher dialogic pairs taken from the National Center for Teacher Effectiveness Main Study (NCTE)\footnote{\url{https://doi.org/10.3886/ICPSR36095.v3}}, a three-year long observation of mathematics instruction. First, they collected human evaluations of \emph{conversational uptake}, a measure of how well the teacher's reply expanded on the student's utterance (e.g., by acknowledging, reformulating, elaborating), as illustrated below.
\begin{center}
\begin{tabularx}{\linewidth}{lX}
\textbf{Student:} & Seven plus seven is fourteen.\\
\textbf{Teacher:} & Okay, so you doubled.  You did your doubles first.  Okay. Fourteen plus eight?\\
 & (\emph{Uptake =  high})\\
\end{tabularx}
\end{center}
Besides human evaluations of uptake, \citet{demszky_measuring_2021} also developed an automated method that could predict uptake as a next-utterance classification task. They fine-tuned a BERT language model \cite{devlin_bert_2019} and found a significant correlation (\(\rho=.54\)) with human evaluations.

This automated measure of conversational uptake can serve as a solid baseline for our study. First, the next-utterance classification predicts uptake based on the preceding student utterance and, therefore, meets our first desideratum. Second, conversational uptake also somehow measures whether a speaker understands the interlocutor. If a teacher's response strongly expands on the student's utterance (i.e., high uptake), it can be deduced that the teacher was able to understand the student. As such, it measures one of the three pedagogical abilities targeted in our second desideratum.
Finally, because we can run the predictive model on different responses to the same student utterance and compare these responses in terms of uptake, the measure meets our third and final desideratum.

\section{Our AI Teacher Test}
\label{sec:org5e573f9}

As a possible solution to the AI teacher challenge described in \Cref{sec--ai-teacher-test-challenge}, we adopted the following method.
First, we ran Blender and GPT-3 on real-world educational dialogues and simulated responses to student utterances. We then paid human raters to compare pairs of responses on several different pedagogical dimensions. Finally, we ran a probabilistic model to compute aggregate scores. In addition, we also ran the model developed by \citet{demszky_measuring_2021} on our data in order to compare our scores to predictions of uptake.

\subsection{Student-Teacher Dialogues}
\label{sec--student-teacher-dialogues}
\begin{table}[h]
\caption{Datasets of Student-Teacher Interactions}
\label{tab--student-teacher-dialogue-data}
\centering
\begin{tabular}{llrr}
\hline
Domain & Dataset & Dialogues & Dialogic Pairs\\
\hline
Language & TSCC \cite{caines_teacherstudent_2020} & 102 & 4439\\
Mathematics & Uptake \cite{demszky_measuring_2021} & 0 & 2246\\
\hline
\end{tabular}
\end{table}

The two datasets used in this study are listed in \Cref{tab--student-teacher-dialogue-data}. The \textbf{\emph{Educational Uptake Dataset}} compiled by \citet{demszky_measuring_2021} includes 2,246 dialogic pairs sampled from the NCTE transcripts (see \Cref{sec--related-work}). The complete dialogue transcripts, however, have not yet been made available. The \textbf{\emph{Teacher-Student Chatroom Corpus} (TSCC)} compiled by \citet{caines_teacherstudent_2020} includes 102 anonymized student-teacher dialogues in second language education.
Each chatroom is a lesson where a teacher converses with a student in order to work on a language exercise and assess the student's English language proficiency. The corpus includes 13,215 turns and 130 turns on average per dialogue. Each utterance is annotated with several metadata, including conversational organization (e.g., opening, closing, eliciting, scaffolding, and revision) and teaching focus (e.g., vocabulary). \Cref{fig--example-dialogue-ability} shows an example excerpt of a teacher's eliciting, scaffolding, and revision.
It should be noted, however, that the TSCC dialogues include many consecutive utterances by either the student or the teacher. Therefore, the data were slightly adapted for this study: all successive utterances by the same speaker were concatenated into one turn such that each conversation was composed of alternating dialogic pairs. As a result, the data included 4,439 student-teacher pairs.

\subsection{Simulating Agent Responses}
\label{sec--conversational-agents}
For each dialogic pair in the student-teacher dialogues, we automatically generated AI teacher responses. We used the ParlAI framework \cite{miller-etal-2017-parlai}
to load the student-teacher dialogues, to generate responses to each student utterance, and to compute several standard evaluation metrics. In this study, we focused on two models. We ran several Blender models (90M, 400M, 3B, 9B parameters) on the language (TSCC) and mathematics (Uptake) educational dialogues. We implemented a new agent that made requests to the OpenAI API in order to obtain generated responses for each student utterance. Each request included a mandatory prompt with instructions for GPT-3 (\emph{The following is a conversation with a teacher. The teacher is polite, helpful, professional, on topic, and factually correct.}), the preceding dialogue history (restricted to meet the maximum number of tokens per request), and the student's utterance. We obtained completions from the smallest (Ada) and largest (Davinci) models.

\subsection{Measuring Pedagogical Ability}
\label{sec:org903f4f1}

\begin{figure}[t]
\centering

\includegraphics[width=\linewidth]{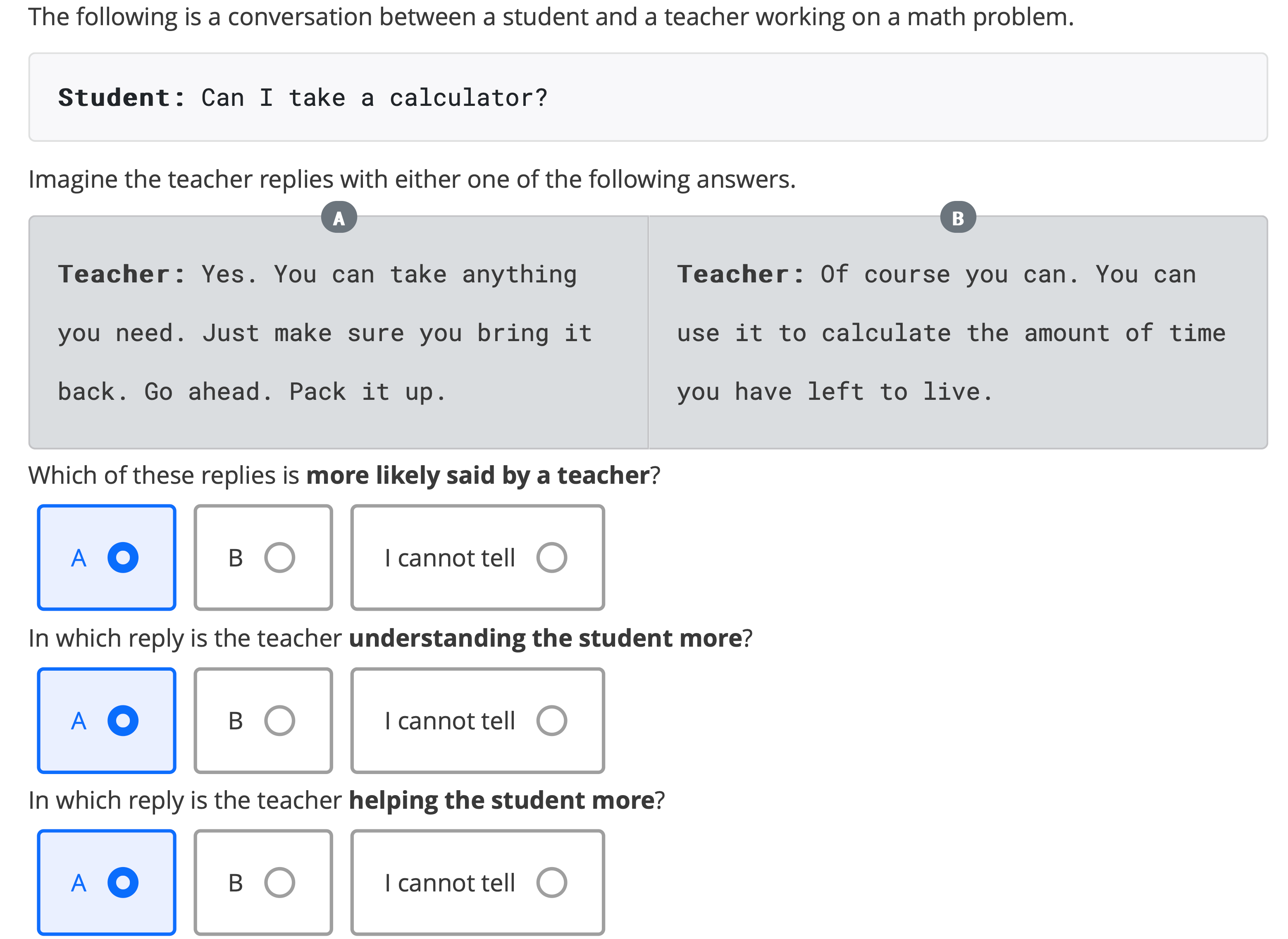}
\caption{Screenshot of the Comparative Judgment Task}
\label{fig--appendix-survey-example-1}
\end{figure}

After collecting AI teacher responses in educational dialogues, we collected evaluations of pedagogical ability via an online survey. First, participants were given a short introduction and a consent form. Then, participants were given the same example to familiarize themselves with the task at hand. In the following comparative judgment task (\Cref{fig--appendix-survey-example-1}), 15 items were randomly and evenly selected from a pool of relevant items. Each item had three components: a dialogue context (limited to 100 tokens), one comparison of two teacher replies, and three questions targeting a pedagogical ability (speak like a teacher, understand the student, and help the student). For each participant, one pairwise comparison was randomly selected from three possible combinations (Teacher vs. Blender, Teacher vs. GPT-3, or Blender vs. GPT-3) and the order of the comparative pair was randomly shuffled.

\textbf{\emph{Item Selection}.}
A crucial challenge in the evaluation process was to pinpoint those teacher utterances that were important to evaluate. In the student-teacher dialogues described in \Cref{sec--student-teacher-dialogues}, not all teacher utterances were necessarily relevant. In fact, many conversational turns were not pertaining to any educational goal, such as opening sequences, closing sequences, and other chit-chat. From the 6,685 eligible dialogic pairs, only those utterances were selected where the teacher was actually eliciting and scaffolding the student's understanding. Additionally, short utterances that comprised of single words or sentence fragments (e.g., \emph{Perfect!}, \emph{Yay!}) were also excluded.\footnote{It should be noted that this exclusion criterion did not apply to the generated responses. As shown in \Cref{fig--example-dialogue-ability}, some generated responses comprised of single words or sentence fragments (e.g., \emph{yes!}). Although this could be seen as giving an advantage to the teacher responses, it was only meant to focus our test on more expressive teacher language. In a future study, we might try to capture the full range of teacher language, from single words to complex utterances.}
Furthermore, the results of a pilot study with eight evaluators highlighted that the dialogic pairs taken from the Uptake dataset were difficult to evaluate because there was no informative context.
Consequently, we focused only on the TSCC dataset for the comparative judgment task, carefully screened the corpus for relevant items and informative dialogue contexts, and ended up with a sample of 52 items.

\textbf{\emph{Participants}.}
We recruited a sample of 120 participants from Prolific Academic, a crowdsourcing platform developed at Oxford University. Participants were prescreened to ensure a balanced gender representation (50\% female, 50\% male).
Study participants were aged 19 to 66 (\(M=33\), \(SD=11.3\); female: \(M=32.4\), \(SD=10.9\); male: \(M=33.5\), \(SD=11.7\)) and resided in the United Kingdom (\(n=86\)) or the United States (\(n=34\)).
On average, participants had an excellent Prolific score of 99.2\% (\(SD=1.4\); female: \(M=99.1\), \(SD=1.6\); male: \(M=99.3\), \(SD=1.3\)) and took 18 minutes to complete the survey (\(SD = 11.2\); female: \(M = 18.9\), \(SD = 11.1\); male: \(M= 17.3\), \(SD = 11.4\)). Because the tasks required a fair amount of cognitive involvement (reading the dialogue, reading different replies, comparing different options), we estimated that the survey would take about 30 minutes. We then used the default payment rate of £7.50/h. Participants were paid according to estimated study completion time (£3.75 for 30 minutes).

\textbf{\emph{Agreement}.} There was a high observed agreement between evaluators on the example given before the comparative judgment task (\Cref{fig--appendix-survey-example-1}). Most agreed that option A (the true teacher response) was \emph{more likely said by a teacher} (95\%), \emph{understanding the student more} (83\%), and \emph{helping the student more} (86\%).

\textbf{\emph{Outlier Detection}.}
To detect potential outliers among the evaluators, we identified those who consistently chose option A or B in the paired comparisons. This first-position (or ``home-field'' advantage) effect was detected by estimating an intercept parameter \(\alpha_0\) in the model described below.
However, instead of estimating different \(\alpha\) parameters for each teacher response (combining the scores of all evaluators), we reversed the method and computed different \(\alpha\) parameters for each evaluator (combining the scores for all items evaluated by the evaluator). Evaluators were excluded when the credible interval around the intercept was above or below zero, which indicated that they were biased towards selecting either option A (CI above zero) or option B (CI below zero). Based on this outlier detection method, the data from seven evaluators were removed.
The remaining data included 4,782 comparisons from 113 evaluators and 10.9 evaluations on average for each pair (Teacher vs. Blender, Teacher vs. GPT-3, or Blender vs. GPT-3).

\textbf{\emph{Bayesian Bradley-Terry Model}.}
A Bradley-Terry model \cite{bradley_rank_1952} is a probabilistic model that predicts the outcome of one or more pairwise comparisons.
Consider \(n\) items (i.e, a student utterance and preceding dialogue), a set of \(t\) possible responses (i.e., Teacher, Blender, GPT-3) to each item, and a set of \(m\) abilities (i.e., speak like a teacher, understand the student, help the student). For each item \(l \in [n]\) and for each ability \(k \in [m]\), we inferred a latent parameter \(\alpha_{ikl}\) for each possible teacher response \(i \in [t]\). The outcome \(y_{ijkl}\) was an independent Bernoulli variable with a parameter \(p_{ijkl} \in [0, 1]\) measuring the chance that, for an item \(l\) and an ability \(k\), teacher response \(i\) would be preferred over teacher response \(j\), for all \(i, j \in [t]\) and \(i \neq j\). This probability was defined as
\begin{align}
p_{ijkl} &:= \sigma\left( \alpha_{ikl} - \alpha_{jkl} \right) \Rightarrow \log \frac{p_{ijkl}}{1-p_{ijkl}} = \alpha_{ikl} - \alpha_{jkl}
\end{align}
where \(\sigma\) is the logistic function \(\sigma(x) = \frac{1}{1+e^{-x}}\) and \(\alpha_i, \alpha_j\) are the latent parameters that measure the strengths of \(i\) and \(j\) respectively. In case of ties (the \emph{I cannot tell} option), the outcome was picked uniformly at random. We used an extended version of the basic Bradley-Terry model including an intercept parameter \(\alpha_0 \in \mathbb{R}\), which measures a ``home-field'' advantage.
\begin{align}
p_{ijkl} := \sigma(\alpha_{0kl} + \alpha_{ikl} - \alpha_{jkl})
\end{align}
If \(\alpha_0 > 0\), there was a greater chance that the evaluator would pick the first element in the comparison. If    \(\alpha_0 = 0\), there was no order effect.
To infer the latent parameters \(\vec{\alpha}_{kl} = (\alpha_{0kl},...,\alpha_{tkl})\), we adopted a Bayesian approach by drawing samples from the posterior \(p(\alpha|y) \propto p(y|\alpha) p(\alpha)\) with a non-conjugate prior distribution, \(\alpha \sim \mathcal{N}(0, 1)\). We used Stan \cite{standevelopmentteam_stan_2022,riddell_pystan_2021} to compute posterior means and 95\% HDI (Highest Density Interval) credible intervals from 4,000 simulations using Hamiltonian Monte Carlo (HMC) sampling \cite{duane_hybrid_1987} and the NUTS (No-U-Turn Sampler) algorithm \cite{homan_nouturn_2014}.
For each simulation, the estimated ability parameters were used to rank each response on each item and for each ability.

\section{Results}
\label{sec:org481fbd4}

\subsection{Baseline: Conversational Uptake}
\label{sec:org22d8994}

\begin{figure}[t]
\centering

\includegraphics[width=\linewidth]{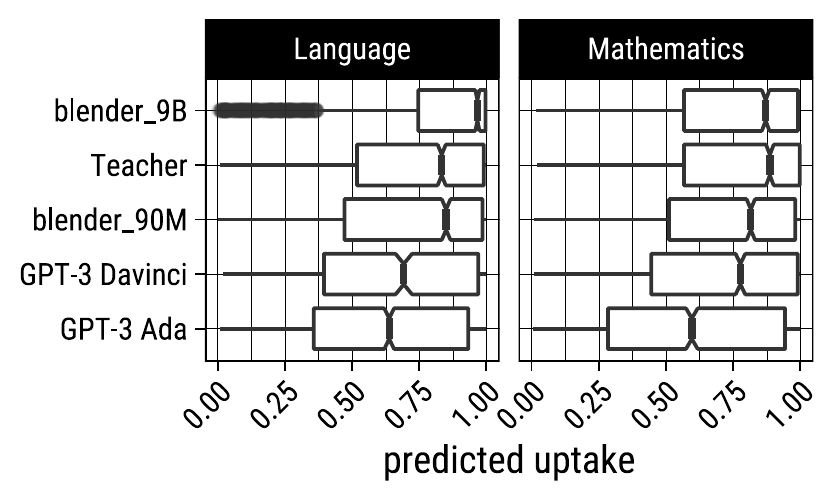}
\caption{Predicted Uptake of Human and AI Teacher Responses in Language and Math Educational Dialogues}
\label{fig--uptake-tscc-uptake-blender-gpt3}
\end{figure}

We start our analyses with a comparison of conversational uptake in human and AI teacher responses, for the two student-teacher dialogue datasets presented in \Cref{sec--student-teacher-dialogues}.
\Cref{fig--uptake-tscc-uptake-blender-gpt3} shows the predicted uptake for the smallest and largest Blender and GPT-3 models, compared to the actual teacher's responses.
The results show that the largest Blender model (with 9B parameters) outperformed all others for both the language (TSCC) and mathematics (Uptake) educational dialogues. This suggests that Blender tended to generate better next utterances to student utterances.

\begin{figure}[t]
\centering

\includegraphics[width=\linewidth]{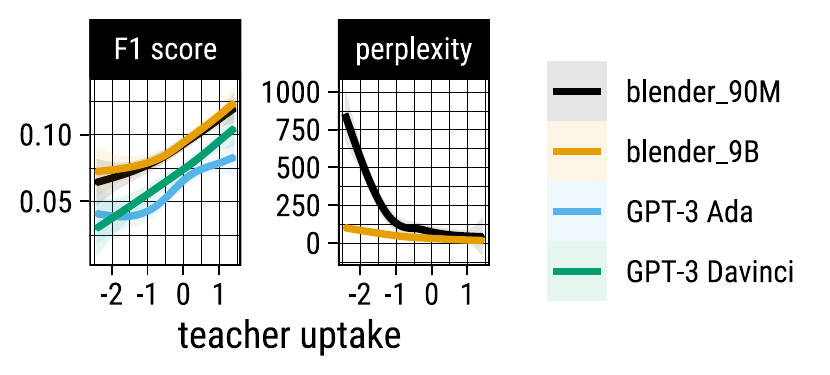}
\caption{Associations Between Generative Performance (Model Perplexity, F1 Unigram Overlap) and True Teacher Uptake (Z-Score) in Mathematics Educational Dialogues}
\label{fig--uptake-perplexity-f1}
\end{figure}

\Cref{fig--uptake-perplexity-f1} zooms in on the AI teacher responses in the mathematics educational dialogues. Several correlation analyses were run to examine the association between generative performance (perplexity and F1 score) and the human annotations of teacher uptake collected by \citet{demszky_measuring_2021}.
Perplexity (lower is better) indicates how well the model can generate a linguistic utterance from its probability distribution, whereas F1 score (higher is better) indicates the unigram overlap between the generated response and the teacher's response.
There was a negative, statistically significant, and large correlation between model perplexity and true teacher uptake, as measured by Pearson's product-moment correlation coefficient, \(r = -0.31\), 95\% CI [-0.34, -0.26], \(t(1996) = -14.32\), \(p < .001\). Similarly, there was a positive, statistically significant, and small correlation between F1 unigram overlap and true teacher uptake, \(r = 0.16\), 95\% CI [0.12, 0.20], \(t(1996) = 7.35\), \(p < .001\).
In other words, Blender tended to generate better responses in cases where the actual teacher was also judged to have given a better response (more uptake). Moreover, this association between generative performance and teacher uptake was observed for all Blender and GPT-3 models (see \Cref{fig--uptake-perplexity-f1}). These findings suggest that some student utterances may be simply easier to reply to, for both human and AI teachers.

\subsection{Our Test: Pedagogical Ability}
\label{sec:orga5f5ccb}

\begin{figure}[t]
\centering

\includegraphics[width=\linewidth]{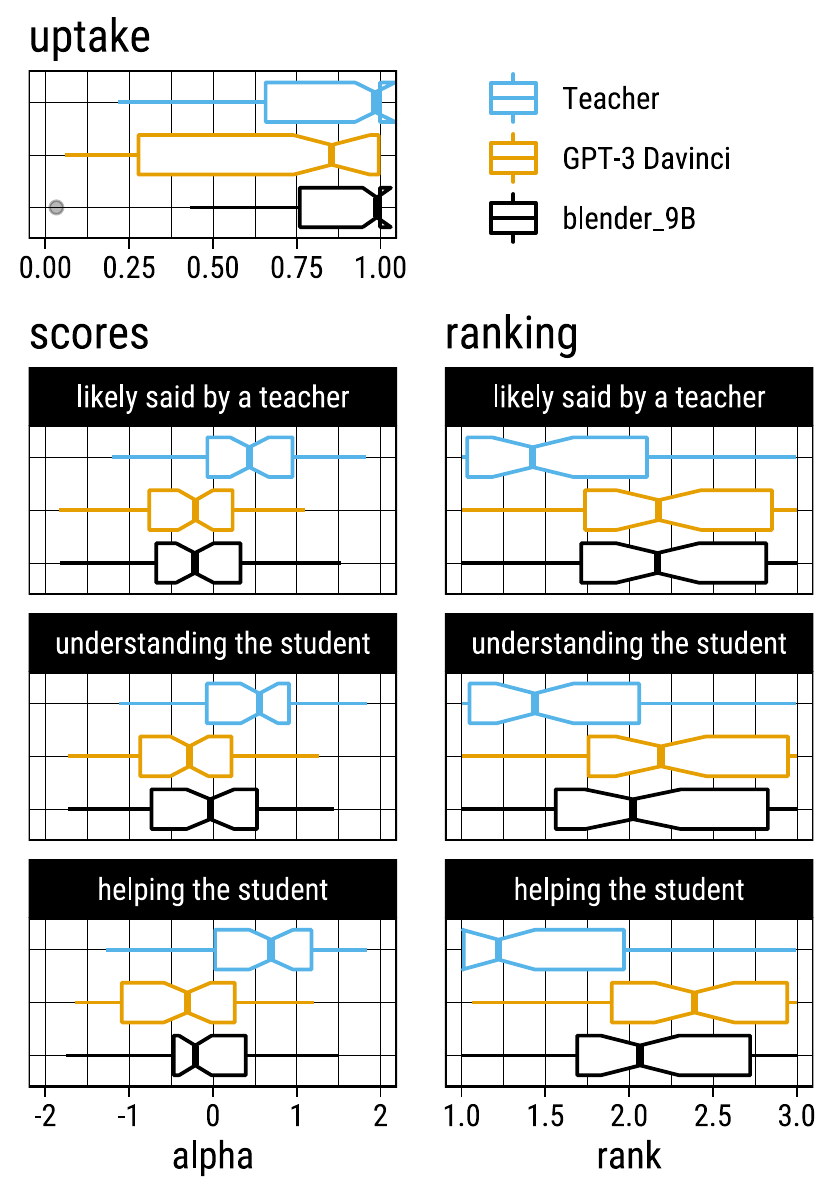}
\caption{Bayesian Estimates and Rankings of Pedagogical Ability in Replying to a Student in an Educational Dialogue, Compared to Predictions of Conversational Uptake}
\label{fig--uptake-alpha-rank-teacher-blenderbot-gpt3-boxplot}
\end{figure}

We now focus all following analyses on the selection of teacher responses that were compared in terms of pedagogical ability.
\Cref{fig--uptake-alpha-rank-teacher-blenderbot-gpt3-boxplot} shows a boxplot of the expected values of \(\alpha\) (and associated rankings) for each possible response to a student utterance on the three pedagogical dimensions. The figure also compares these scores to predictions of conversational uptake.
In terms of conversational uptake, the results showed no significant differences between human and AI teachers, as indicated by the overlapping notches in the boxplot. In terms of pedagogical ability, however, a one-way ANOVA revealed a statistically significant difference between human teachers and AI teachers on the three dimensions cited above, \(F(2,144)=13.1\), \(p<.001\), \(F(2,144)=11.8\), \(p<.001\), \(F(2,144)=22.3\), \(p<.001\), respectively.\footnote{A Shapiro-Wilk test showed that the assumption of normality was not violated for any of the three pedagogical dimensions,   \(W=0.99\), \(p=0.76\), \(W=0.99\), \(p=0.19\), \(W=0.99\), \(p=0.15\), respectively.}
Tukey's HSD post hoc test for multiple comparisons showed that, compared to the actual teacher, the mean ability of Blender was significantly lower for speaking like a teacher (\(\Delta \, \text{alpha} = -0.60\), [95CI \(-0.93\), \(-0.26\)], \(p<.001\)), understanding the student (\(\Delta \, \text{alpha} = -0.55\), [95CI \(-0.90\), \(-0.20\)], \(p<.001\)), and helping the student (\(\Delta \, \text{alpha} = -0.75\), [95CI \(-1.10\), \(-0.40\)], \(p<.001\)).
Similarly, compared to the actual teacher, the mean ability of GPT-3 was significantly lower for speaking like a teacher (\(\Delta \, \text{alpha} = -0.67\), [95CI \(-1.00\), \(-0.33\)], \(p<.001\)), understanding the student (\(\Delta \, \text{alpha} = -0.67\), [95CI \(-1.02\), \(-0.32\)], \(p<.001\)), and helping the student (\(\Delta \, \text{alpha} = -0.93\), [95CI \(-1.28\), \(-0.58\)], \(p<.001\)).
As for Blender and GPT-3, there was no statistically significant difference between the two when it came to speaking like a teacher (\(\Delta \, \text{alpha} = 0.071\), \(p=.41\)). Overall, Blender seemed better at understanding (\(\Delta \, \text{alpha} = +0.12\)) and helping (\(\Delta \, \text{alpha} = +0.18\)) the student but these differences were not significant (\(p=.47\), \(p=.53\), respectively).

\Cref{tab--pearson-correlations-uptake-ability} shows that our estimates of pedagogical ability were significantly correlated with conversational uptake. Interestingly, the highest correlation was observed for the ability of understanding the student. This result was not surprising: as previously noted in \Cref{sec--related-work}, uptake also somehow measures whether a speaker understands the interlocutor.

\begin{table}[t]
\caption{Pearson Correlations Between Uptake and Ability}
\label{tab--pearson-correlations-uptake-ability}
\centering
\begin{tabularx}{\linewidth}{|X|r|r|r|r|}
\hline
 & \(r\) & \(t\) & df & \(p\)\\
\hline
likely said by a teacher & .35 & 3.47 & 85 & <.001\\
understanding the student & \textbf{.38} & 3.82 & 85 & <.001\\
helping the student & .33 & 3.27 & 85 & .002\\
\hline
\end{tabularx}
\end{table}

\begin{table}[t]
\caption{Percentage of Replies with a Positive Ability or where the 95\% CI Excludes Zero (Either Above or Below)}
\label{tab--percentages-positive-alpha-zero-notin-ci}
\centering
\begin{tabularx}{\linewidth}{|l|X|r|r|}
\hline
Agent & Ability & \(\alpha > 0\) & \(0 \not\in\) CI\\
\hline
Teacher & speak like a teacher & 69\% & 8\%\\
Teacher & understand the student & 71\% & 6\%\\
Teacher & \textbf{help the student} & \textbf{78\%} & \textbf{14\%}\\
\hline
Blender 9B & speak like a teacher & 41\% & 6\%\\
Blender 9B & understand the student & 45\% & 10\%\\
Blender 9B & help the student & 35\% & 8\%\\
\hline
GPT-3 Dav. & speak like a teacher & 35\% & 6\%\\
GPT-3 Dav. & understand the student & 35\% & 2\%\\
GPT-3 Dav. & help the student & 33\% & 12\%\\
\hline
\end{tabularx}
\end{table}

Finally, we examined the pedagogical ability of human and AI teachers on a per-item basis.
\Cref{tab--percentages-positive-alpha-zero-notin-ci} shows the percentage of teacher replies for which the expected value of \(\alpha\) was positive. For 69-78\% of dialogue contexts, the actual teacher's response was positively evaluated and, therefore, highly ranked.
For helpfulness in particular, 78\% of human teacher responses were positively evaluated. There was also a higher percentage (14\%) of certainty in the estimates of helpfulness.
Nevertheless, there were many cases (22-31\% of dialogue contexts) where the actual teacher's response was negatively evaluated as well as cases (33-45\%) where artificially generated responses were positively evaluated. The following example illustrates a case where the response generated by Blender was largely preferred over the actual teacher's response.

\begin{center}
\begin{tabularx}{\linewidth}{|lX|}
\hline
\textbf{Teacher:} & People are looking forward to it = there is a lot of \ldots{} ?\\
\textbf{Student:} & excitement\\
\hline
\#1 \textbf{Blender:} & Excitement is a good one! People are very excited about it. Do you have any more?\\
 & - \emph{likely said by a teacher = 1.52}\\
 & - \emph{understanding the student = 1.32}\\
 & - \emph{helping the student = 1.52}\\
\hline
\#2 \textbf{Teacher:} & ok good! Coronavirus has caused a lot of \ldots{} among the population.\\
 & - \emph{likely said by a teacher = -0.90}\\
 & - \emph{understanding the student = -0.57}\\
 & - \emph{helping the student = -0.61}\\
\hline
\#3 \textbf{GPT-3:} & excitement about the next World Cup\ldots{}\\
 & - \emph{likely said by a teacher = -0.61}\\
 & - \emph{understanding the student =  -0.60}\\
 & - \emph{helping the student = -0.90}\\
\hline
\end{tabularx}
\end{center}

\section{Concluding Discussion}
\label{sec:org995925f}

How well are state-of-the-art conversational agents, such as Blender and GPT-3, capable of replying to a student in an educational dialogue?
When it comes to uptaking from and expanding on a student's utterance, Blender comes out on top, outperforming the actual teacher and GPT-3. Based on the results of our AI teacher test, we come to similar conclusions. Although our test does not corroborate that Blender can actually outperform a human teacher, there is nevertheless a closer gap with human performance when it comes to understanding the student. Blender scores noticeably better on this specific pedagogical dimension, with a higher percentage of positively evaluated responses. These findings may be attributed to Blender's particular training objective, namely blended skill talk. By learning to be more empathetic, Blender might be incidentally learning to take up more from and be more understanding of its interlocutor. By contrast, the results of our AI teacher test show that GPT-3 performs quantifiably worse than Blender and significantly worse than real teachers on all measured abilities, despite its proven capacity for few-shot in-context learning. What is more, both Blender and GPT-3 are well behind human performance when it comes to helping the student.

A secondary finding of our AI teacher test is that not all human teacher responses are necessarily positively evaluated. Even though the AI teacher responses generally fall short regarding pedagogical ability, we could still leverage generated responses as a means of sampling and recommending potentially better responses.

The solution proposed in this paper is surely not a perfect test, but it is a first step towards building much-needed evaluation methods.

\section{Acknowledgments}
\label{sec:orgb6579b4}

This research was supported by a fellowship of the Belgian American Educational Foundation (to the first author) and by a grant from Stanford HAI. We thank Andrew Caines, Dora Demszky, Noah Goodman, and our colleagues for their valuable help and suggestions. We thank all anonymous reviewers for their insights that improved the paper.

\clearpage

\bibliographystyle{abbrvnat}
\bibliography{main}

\begin{thebibliography}{20}
\providecommand{\natexlab}[1]{#1}
\providecommand{\url}[1]{\texttt{#1}}
\expandafter\ifx\csname urlstyle\endcsname\relax
  \providecommand{\doi}[1]{doi: #1}\else
  \providecommand{\doi}{doi: \begingroup \urlstyle{rm}\Url}\fi

\bibitem[Bibauw et~al.(2022)Bibauw, {Van den Noortgate}, Fran{\c c}ois, and
  Desmet]{bibauw_dialogue_2022}
S.~Bibauw, W.~{Van den Noortgate}, T.~Fran{\c c}ois, and P.~Desmet.
\newblock Dialogue systems for language learning: A meta-analysis.
\newblock \emph{Language Learning \& Technology}, 26\penalty0 (1):\penalty0
  accepted, 2022.

\bibitem[Bommasani et~al.(2021)Bommasani, Hudson, Adeli, Altman, Arora, {von
  Arx}, Bernstein, Bohg, Bosselut, Brunskill, Brynjolfsson, Buch, Card,
  Castellon, Chatterji, Chen, Creel, Davis, Demszky, Donahue, Doumbouya,
  Durmus, Ermon, Etchemendy, Ethayarajh, {Fei-Fei}, Finn, Gale, Gillespie,
  Goel, Goodman, Grossman, Guha, Hashimoto, Henderson, Hewitt, Ho, Hong, Hsu,
  Huang, Icard, Jain, Jurafsky, Kalluri, Karamcheti, Keeling, Khani, Khattab,
  Kohd, Krass, Krishna, Kuditipudi, Kumar, Ladhak, Lee, Lee, Leskovec, Levent,
  Li, Li, Ma, Malik, Manning, Mirchandani, Mitchell, Munyikwa, Nair, Narayan,
  Narayanan, Newman, Nie, Niebles, Nilforoshan, Nyarko, Ogut, Orr,
  Papadimitriou, Park, Piech, Portelance, Potts, Raghunathan, Reich, Ren, Rong,
  Roohani, Ruiz, Ryan, R{\'e}, Sadigh, Sagawa, Santhanam, Shih, Srinivasan,
  Tamkin, Taori, Thomas, Tram{\`e}r, Wang, Wang, Wu, Wu, Wu, Xie, Yasunaga,
  You, Zaharia, Zhang, Zhang, Zhang, Zhang, Zheng, Zhou, and
  Liang]{bommasani_opportunities_2021}
R.~Bommasani, D.~A. Hudson, E.~Adeli, R.~Altman, S.~Arora, S.~{von Arx}, M.~S.
  Bernstein, J.~Bohg, A.~Bosselut, E.~Brunskill, E.~Brynjolfsson, S.~Buch,
  D.~Card, R.~Castellon, N.~Chatterji, A.~Chen, K.~Creel, J.~Q. Davis,
  D.~Demszky, C.~Donahue, M.~Doumbouya, E.~Durmus, S.~Ermon, J.~Etchemendy,
  K.~Ethayarajh, L.~{Fei-Fei}, C.~Finn, T.~Gale, L.~Gillespie, K.~Goel,
  N.~Goodman, S.~Grossman, N.~Guha, T.~Hashimoto, P.~Henderson, J.~Hewitt,
  D.~E. Ho, J.~Hong, K.~Hsu, J.~Huang, T.~Icard, S.~Jain, D.~Jurafsky,
  P.~Kalluri, S.~Karamcheti, G.~Keeling, F.~Khani, O.~Khattab, P.~W. Kohd,
  M.~Krass, R.~Krishna, R.~Kuditipudi, A.~Kumar, F.~Ladhak, M.~Lee, T.~Lee,
  J.~Leskovec, I.~Levent, X.~L. Li, X.~Li, T.~Ma, A.~Malik, C.~D. Manning,
  S.~Mirchandani, E.~Mitchell, Z.~Munyikwa, S.~Nair, A.~Narayan, D.~Narayanan,
  B.~Newman, A.~Nie, J.~C. Niebles, H.~Nilforoshan, J.~Nyarko, G.~Ogut, L.~Orr,
  I.~Papadimitriou, J.~S. Park, C.~Piech, E.~Portelance, C.~Potts,
  A.~Raghunathan, R.~Reich, H.~Ren, F.~Rong, Y.~Roohani, C.~Ruiz, J.~Ryan,
  C.~R{\'e}, D.~Sadigh, S.~Sagawa, K.~Santhanam, A.~Shih, K.~Srinivasan,
  A.~Tamkin, R.~Taori, A.~W. Thomas, F.~Tram{\`e}r, R.~E. Wang, W.~Wang, B.~Wu,
  J.~Wu, Y.~Wu, S.~M. Xie, M.~Yasunaga, J.~You, M.~Zaharia, M.~Zhang, T.~Zhang,
  X.~Zhang, Y.~Zhang, L.~Zheng, K.~Zhou, and P.~Liang.
\newblock On the {{Opportunities}} and {{Risks}} of {{Foundation Models}}.
\newblock Technical report, {Stanford University}, {Center for Research on
  Foundation Models (CRFM)}, Aug. 2021.

\bibitem[Bradley and Terry(1952)]{bradley_rank_1952}
R.~A. Bradley and M.~E. Terry.
\newblock Rank {{Analysis}} of {{Incomplete Block Designs}}: {{I}}. {{The
  Method}} of {{Paired Comparisons}}.
\newblock \emph{Biometrika}, 39\penalty0 (3/4):\penalty0 324, Dec. 1952.
\newblock ISSN 00063444.
\newblock \doi{10.2307/2334029}.

\bibitem[Brown et~al.(2020)Brown, Mann, Ryder, Subbiah, Kaplan, Dhariwal,
  Neelakantan, Shyam, Sastry, Askell, Agarwal, {Herbert-Voss}, Krueger,
  Henighan, Child, Ramesh, Ziegler, Wu, Winter, Hesse, Chen, Sigler, Litwin,
  Gray, Chess, Clark, Berner, McCandlish, Radford, Sutskever, and
  Amodei]{brown_language_2020}
T.~Brown, B.~Mann, N.~Ryder, M.~Subbiah, J.~D. Kaplan, P.~Dhariwal,
  A.~Neelakantan, P.~Shyam, G.~Sastry, A.~Askell, S.~Agarwal,
  A.~{Herbert-Voss}, G.~Krueger, T.~Henighan, R.~Child, A.~Ramesh, D.~Ziegler,
  J.~Wu, C.~Winter, C.~Hesse, M.~Chen, E.~Sigler, M.~Litwin, S.~Gray, B.~Chess,
  J.~Clark, C.~Berner, S.~McCandlish, A.~Radford, I.~Sutskever, and D.~Amodei.
\newblock Language models are few-shot learners.
\newblock In H.~Larochelle, M.~Ranzato, R.~Hadsell, M.~F. Balcan, and H.~Lin,
  editors, \emph{Advances in Neural Information Processing Systems}, volume~33,
  pages 1877--1901. {Curran Associates, Inc.}, 2020.

\bibitem[Caines et~al.(2020)Caines, Yannakoudakis, Edmondson, Allen,
  {P{\'e}rez-Paredes}, Byrne, and Buttery]{caines_teacherstudent_2020}
A.~Caines, H.~Yannakoudakis, H.~Edmondson, H.~Allen, P.~{P{\'e}rez-Paredes},
  B.~Byrne, and P.~Buttery.
\newblock The teacher-student chatroom corpus.
\newblock In \emph{Proceedings of the 9th Workshop on {{NLP}} for Computer
  Assisted Language Learning}, pages 10--20, {Gothenburg, Sweden}, Nov. 2020.
  {LiU Electronic Press}.

\bibitem[Demszky et~al.(2021)Demszky, Liu, Mancenido, Cohen, Hill, Jurafsky,
  and Hashimoto]{demszky_measuring_2021}
D.~Demszky, J.~Liu, Z.~Mancenido, J.~Cohen, H.~Hill, D.~Jurafsky, and
  T.~Hashimoto.
\newblock Measuring {{Conversational Uptake}}: {{A Case Study}} on
  {{Student-Teacher Interactions}}.
\newblock In \emph{Proceedings of the 59th {{Annual Meeting}} of the
  {{Association}} for {{Computational Linguistics}} and the 11th
  {{International Joint Conference}} on {{Natural Language Processing}}
  ({{Volume}} 1: {{Long Papers}})}, pages 1638--1653, {Online}, 2021.
  {Association for Computational Linguistics}.
\newblock \doi{10.18653/v1/2021.acl-long.130}.

\bibitem[Devlin et~al.(2019)Devlin, Chang, Lee, and
  Toutanova]{devlin_bert_2019}
J.~Devlin, M.-W. Chang, K.~Lee, and K.~Toutanova.
\newblock {{BERT}}: {{Pre-training}} of deep bidirectional transformers for
  language understanding.
\newblock In \emph{Proceedings of the 2019 Conference of the North {{American}}
  Chapter of the Association for Computational Linguistics: {{Human}} Language
  Technologies, Volume 1 (Long and Short Papers)}, pages 4171--4186,
  {Minneapolis, Minnesota}, June 2019. {Association for Computational
  Linguistics}.
\newblock \doi{10.18653/v1/N19-1423}.

\bibitem[Duane et~al.(1987)Duane, Kennedy, Pendleton, and
  Roweth]{duane_hybrid_1987}
S.~Duane, A.~D. Kennedy, B.~J. Pendleton, and D.~Roweth.
\newblock Hybrid {{Monte Carlo}}.
\newblock \emph{Physics Letters B}, 195\penalty0 (2):\penalty0 216--222, Sept.
  1987.
\newblock ISSN 0370-2693.
\newblock \doi{10.1016/0370-2693(87)91197-X}.

\bibitem[Goe et~al.(2008)Goe, Bell, and Little]{goe_approaches_2008}
L.~Goe, C.~Bell, and O.~Little.
\newblock \emph{Approaches to {{Evaluating Teacher Effectiveness}}: {{A
  Research Synthesis}}}.
\newblock {National Comprehensive Center for Teacher Quality}, June 2008.

\bibitem[Heldsinger and Humphry(2010)]{heldsinger_using_2010}
S.~Heldsinger and S.~Humphry.
\newblock Using the {{Method}} of {{Pairwise Comparison}} to {{Obtain Reliable
  Teacher Assessments}}.
\newblock \emph{Australian Educational Researcher}, 37\penalty0 (2):\penalty0
  1--19, Aug. 2010.
\newblock ISSN 0311-6999.

\bibitem[Homan and Gelman(2014)]{homan_nouturn_2014}
M.~D. Homan and A.~Gelman.
\newblock The {{No-U-turn}} sampler: Adaptively setting path lengths in
  {{Hamiltonian Monte Carlo}}.
\newblock \emph{The Journal of Machine Learning Research}, 15\penalty0
  (1):\penalty0 1593--1623, Jan. 2014.
\newblock ISSN 1532-4435.

\bibitem[Lesterhuis et~al.(2017)Lesterhuis, Verhavert, Coertjens, Donche, and
  De~Maeyer]{lesterhuis_comparative_2017}
M.~Lesterhuis, S.~Verhavert, L.~Coertjens, V.~Donche, and S.~De~Maeyer.
\newblock Comparative judgement as a promising alternative to score
  competences.
\newblock In E.~Cano, G.~Ion, and J.~Keengwe, editors, \emph{Innovative
  {{Practices}} for {{Higher Education Assessment}} and {{Measurement}}:},
  Advances in {{Higher Education}} and {{Professional Development}}. {IGI
  Global}, 2017.
\newblock ISBN 978-1-5225-0531-0 978-1-5225-0532-7.
\newblock \doi{10.4018/978-1-5225-0531-0}.

\bibitem[Miller et~al.(2017)Miller, Feng, Batra, Bordes, Fisch, Lu, Parikh, and
  Weston]{miller-etal-2017-parlai}
A.~Miller, W.~Feng, D.~Batra, A.~Bordes, A.~Fisch, J.~Lu, D.~Parikh, and
  J.~Weston.
\newblock {{ParlAI}}: {{A}} dialog research software platform.
\newblock In \emph{Proceedings of the 2017 Conference on Empirical Methods in
  Natural Language Processing: {{System}} Demonstrations}, pages 79--84,
  {Copenhagen, Denmark}, Sept. 2017. {Association for Computational
  Linguistics}.
\newblock \doi{10.18653/v1/D17-2014}.

\bibitem[Muijs(2006)]{muijs_measuring_2006}
D.~Muijs.
\newblock Measuring teacher effectiveness: {{Some}} methodological reflections.
\newblock \emph{Educational Research and Evaluation}, 12\penalty0 (1):\penalty0
  53--74, Feb. 2006.
\newblock ISSN 1380-3611, 1744-4187.
\newblock \doi{10.1080/13803610500392236}.

\bibitem[Pillutla et~al.(2021)Pillutla, Swayamdipta, Zellers, Thickstun,
  Welleck, Choi, and Harchaoui]{pillutla_mauve_2021}
K.~Pillutla, S.~Swayamdipta, R.~Zellers, J.~Thickstun, S.~Welleck, Y.~Choi, and
  Z.~Harchaoui.
\newblock {{MAUVE}}: {{Measuring}} the {{Gap Between Neural Text}} and {{Human
  Text}} using {{Divergence Frontiers}}.
\newblock In \emph{Advances in {{Neural Information Processing Systems}} 34
  Pre-Proceedings ({{NeurIPS}} 2021)}, pages 1--35, Nov. 2021.

\bibitem[Riddell et~al.(2021)Riddell, Hartikainen, and
  Carter]{riddell_pystan_2021}
A.~Riddell, A.~Hartikainen, and M.~Carter.
\newblock {{PyStan}} (3.3.0).
\newblock PyPI, Mar. 2021.

\bibitem[Roller et~al.(2020)Roller, Dinan, Goyal, Ju, Williamson, Liu, Xu, Ott,
  Shuster, Smith, Boureau, and Weston]{roller_recipes_2020}
S.~Roller, E.~Dinan, N.~Goyal, D.~Ju, M.~Williamson, Y.~Liu, J.~Xu, M.~Ott,
  K.~Shuster, E.~M. Smith, Y.-L. Boureau, and J.~Weston.
\newblock Recipes for building an open-domain chatbot.
\newblock 2020.
\newblock \doi{10.48550/ARXIV.2004.13637}.

\bibitem[Smith et~al.(2020)Smith, Williamson, Shuster, Weston, and
  Boureau]{smith_can_2020}
E.~M. Smith, M.~Williamson, K.~Shuster, J.~Weston, and Y.-L. Boureau.
\newblock Can you put it all together: {{Evaluating}} conversational agents'
  ability to blend skills.
\newblock In \emph{Proceedings of the 58th Annual Meeting of the Association
  for Computational Linguistics}, pages 2021--2030, {Online}, July 2020.
  {Association for Computational Linguistics}.
\newblock \doi{10.18653/v1/2020.acl-main.183}.

\bibitem[{Stan Development Team}(2022)]{standevelopmentteam_stan_2022}
{Stan Development Team}.
\newblock Stan {{Modeling Language Users Guide}} and {{Reference Manual}}
  (v2.29.0), 2022.

\bibitem[Wollny et~al.(2021)Wollny, Schneider, Di~Mitri, Weidlich, Rittberger,
  and Drachsler]{wollny_are_2021}
S.~Wollny, J.~Schneider, D.~Di~Mitri, J.~Weidlich, M.~Rittberger, and
  H.~Drachsler.
\newblock Are {{We There Yet}}? - {{A Systematic Literature Review}} on
  {{Chatbots}} in {{Education}}.
\newblock \emph{Frontiers in Artificial Intelligence}, 4:\penalty0 654924, July
  2021.
\newblock ISSN 2624-8212.
\newblock \doi{10.3389/frai.2021.654924}.

\end{thebibliography}
\end{document}